\documentclass{article}

    \PassOptionsToPackage{numbers}{natbib}

    \usepackage[final]{neurips_2019}
\usepackage[utf8]{inputenc} 
\usepackage[T1]{fontenc}    
\usepackage{hyperref}       
\usepackage{url}            
\usepackage{booktabs}       
\usepackage{amsfonts}       
\usepackage{nicefrac}       
\usepackage{microtype}      

\usepackage{amsmath, amsthm}
\usepackage{bm, bbm} 
\usepackage{subcaption}
\usepackage{graphicx}

\newcommand\independent{\protect\mathpalette{\protect\independenT}{\perp}}
\def\independenT#1#2{\mathrel{\rlap{$#1#2$}\mkern2mu{#1#2}}}
\DeclareMathOperator*{\argmax}{arg\,max}

\title{Optimising Individual-Treatment-Effect Using Bandits}

\author{%
  Jeroen berrevoets\\
  University of Brussels (VUB)\\
  \texttt{jeroen.berrevoets@vub.ac.be}
  \And
  Sam Verboven\\
  University of Brussels (VUB)\\
  \texttt{sam.verboven@vub.ac.be}
  \And
  Wouter Verbeke\\
  University of Brussels (VUB)\\
  \texttt{wouter.verbeke@vub.ac.be}
}

\begin{document}

\maketitle

\begin{abstract}
	Applying causal inference models in areas such as 
	economics, healthcare and marketing receives great 
	interest from the machine learning community. In 
	particular, estimating the individual-treatment-effect 
	(ITE) in settings such as precision medicine and targeted
	advertising has peaked in application. 
	Optimising this ITE under the \textit{strong-ignorability-assumption}
	--- meaning all confounders expressing influence on the outcome 
	of a treatment are registered in the data 
	--- is often referred to as uplift modeling (UM).
	While these techniques have proven useful in many settings, 
	they suffer vividly in a dynamic environment due to concept drift. 
	Take for example the negative influence on a marketing campaign 
	when a competitor product is released. To counter this, we propose 
	the \textit{uplifted contextual multi-armed bandit} (U-CMAB), a novel 
	approach to optimise the ITE by drawing upon bandit literature. 
	Experiments on real and simulated data indicate that our proposed approach 
	compares favourably against the state-of-the-art. All our code can be found online at \url{https://github.com/vub-dl/u-cmab}.
\end{abstract}

\section{Introduction}
	Making individual-level causal predictions is an important problem in many fields. For example, \textit{individual-treatment-effect} (ITE) predictions can be used to: prescribe medicine only when it causes the best outcome for a specific patient; or advertise only to those that were not going to buy otherwise.

	While many ITE prediction methods exist, they fail to adapt through time. We believe this is a crucial problem in causal inference as many environments are dynamic in nature: patients could build a tolerance to their prescribed medicine; or the initial marketing campaign could suffer from a competitor's product release \cite{fang2018}. In machine learning, we refer to deteriorating behaviour due to a changing environment, as concept drift \cite{tsymbal2004,gama2014}. 

	A first naive attempt to create dynamic causal inference models, could be an adapted on-line learning method, e.g., on-line random forests \cite{saffari2009}. However, such methods require a target variable---which is absent as a counterfactual outcome is unobservable. A second naive approach would be to use a change detection algorithm \cite{gama2014}, initiating a retraining subroutine when necessary. In fact, we have done exactly this in our experiments, but found them to perform poorly compared to our method. 

	We take a fundamentally different approach than the naive strategies described above, as we reformulate uplift modeling in a bandit problem \cite{robbins1952}. Since bandits learn continuously, they easily adapt to dynamic environments using a windowed estimation of their target \cite{sutton2018}. 

\section{Preliminaries and Background}
	\textbf{Uplift models} estimate the net impact of a treatment $ T \in \{0, 1\} $ 
	on a response $ Y \in \{0, 1\} $ for an individual $ \mathbf{x} \in \mathbb{R}^n $. 
	Such net impact is measured through an incremental probability:
	$\hat{u}(Y, T, \mathbf{x}) \doteq \hat{p}(Y=1 | T=1, \mathbf{x}) - \hat{p}(Y=1 | T=0, \mathbf{x})$,
	where $ T=1 $ when the treatment is applied and $ T=0 $ when it is not 
	\citep{devriendt2018, gutierrez2017}. Given a high $ \hat{u} $, we derive 
	that an $ \mathbf{x} $ can be \textit{caused} to respond ($ Y=1 $) to the 
	treatment \cite{devriendt2018, rubin2005}. 
	
	Uplift models are then employed to identify a subpopulation with high 
	$ \hat{u} $. By limiting treatment to this subpopulation we reduce 
	over-treatment by refraining from treating individuals indifferent to 
	treatment ($ \hat{u} = 0 $) or worse, individuals that are averse 
	($ \hat{u} < 0 $) to it. 

	Typically, datasets in UM are built using a randomised trial setting, where 
	$Y \independent T|\mathbf{x}$ and $0 < p(T=1|\mathbf{x}) < 1$ for all $ \mathbf{x}$, assuring 
	the strong-ignoreability-assumption \cite{rubin2005,shalit2017,devriendt2018}. Hence, use of the $do$-operator is not 
	required, contrasting the case when strong-ignoreability is violated \cite{pearl2009}.
	
	\textbf{Contextual multi-armed bandits (CMAB)} differ from UM as they apply 
	treatment in function of expected response only. We define this response as 
	$r(T=i, \mathbf{x}) \doteq \mathbb{E}[R(Y) | T=i, \mathbf{x}]$, where: 
	$ \mathbf{x} $ is considered a context; $ \{T=0, T=1\} $ is the set of 
	arms; and $ R: Y \rightarrow \mathbb{R} $ is the numerical reward for $ Y $ 
	\cite{zhou2015, kuleshov2014}. Optimal treatment selection is then motivated by 
	an estimation of this expected response $\hat{r}$ as in (\ref{eq:bellman}),
	\begin{equation}
		T^*_b = \argmax_i\left\{\hat{r}(T=i, \mathbf{x})\right\} \label{eq:bellman}.
	\end{equation}
	The treatment $ T^*_b $ is 
	chosen over other treatments even if $ T^*_b $ offers only a marginally 
	higher expected response.

	This formulation suggests two major components in a CMAB's objective: (i) 
	response estimation through $\hat{r}$; and (ii) proper treatment selection 
	through (\ref{eq:bellman}).
	Randomly applying treatments ensures $\hat{r}$ to be unbiased, but contrasts 
	the second objective. Balancing these components is often referred to as the
	exploration-exploitation trade-off \cite{sutton2018}. We use this formulation
	to frame our experiments in Section~\ref{sec:results}.

	\textbf{The difference between UM and CMABs} is apparent through the maximisation in (\ref{eq:bellman}). 
	Such maximisation contrasts UM as uplift models 
	inform a decision maker to make \textit{causal} decisions, only applying a treatment 
	when the treatment has a \textit{sufficient} positive effect on $ \mathbf{x} $, 
	i.e., when $ \hat{u} $ is higher than some threshold $ \tau \in [-1, 1) $. 
	As such, the optimal treatment in UM is found using,
	\begin{equation} \label{eq:tau}
		T^*_u = \mathbb{I}\left[ \hat{u}(Y, T, \mathbf{x}) > \tau \right],
	\end{equation}
	where $ \mathbb{I}[\cdot] $ is the indicator function. Using our notation, this difference is simply: $T^*_b \neq T^*_u$.

	We contribute by defining $\tau$, indicating when $ \hat{u} $ is
	considered \textit{high enough}. We then apply 
	our findings to bandit algorithms, making them optimise for uplift. 
	By leveraging the ability to learn continuously the U-CMAB 
	offers resilience in a dynamic environment for individual-level 
	causal models.

\section{Model} 
	Introducing a penalty $ \bm{\psi} $ associated with the 
	cost of the treatment --- with $ T=i \rightarrow \psi_i \in \mathbb{R}$ and 
	$ \bm{\psi} = [\psi_0, \psi_1]^\top $ --- enables causal decision 
	making by the U-CMAB. While $ \tau $ is generally chosen heuristically \cite{devriendt2018}, 
	we provide an analytical method based on $ \bm{\psi} $:
	\begin{equation} \label{eq:ucmab}
		\tau = \frac{\psi_1 - \psi_0}{R(Y=1)},
	\end{equation}
	where: $ \psi_1 $ is the penalty of applying the treatment ($ T=1 $);
	$ \psi_0 $ is the penalty of not applying the treatment ($ T=0 $); and
	$ R(Y=1) $ is the potential (numerical) reward when $ \mathbf{x} $
	responds.

	Two benefits of
	(\ref{eq:ucmab}) come to mind: (i) $ \tau $ is now composed of 
	parameters we can share with a bandit algorithm, and (ii)
	there is an intuitive appeal to (\ref{eq:ucmab})---when  $ \psi_1 $ 
	is high, so is $ \tau $, translating in the requirement of a high 
	$ \hat{u} $ before treatment is applied, i.e., before applying an expensive 
	treatment it should have higher net impact when compared to an inexpensive treatment.

	Once $ \bm{\psi} $ is chosen according to (\ref{eq:ucmab}), it is to be 
	deducted from the bandit's estimated reward $ \hat{r}(T, \mathbf{x}) $,
	\begin{equation} \label{eq:rew}
		\hat{r}_u(T=i, \mathbf{x}) \doteq  \mathbb{E}\left[R(Y) - \psi_i | T=i, \mathbf{x} \right],
	\end{equation}
	creating a new form of reward, $\hat{r}_u$, associated with every $T=i$. 
	
	When 
	$ \hat{r} $ is replaced with $ \hat{r}_u $, optimal treatment selection 
	through (\ref{eq:bellman}) will be altered. 
	Operating according to this $\hat{r}_u$ will yield treatment decisions similar 
	to those made by an uplift model respecting some threshold $ \tau $. We back 
	this claim through experiments (in Section~\ref{sec:results}) and a proof of 
	(\ref{eq:ucmab}) in the Appendix.

	Some intuition into (\ref{eq:rew}) can be achieved by formulating a Markov decision process (MDP), 
	$\langle \mathcal{X}, \mathcal{T}, \mathcal{Y}, \mathbbm{t}, R \rangle$, where: $ \mathcal{X} $ is the set of individuals, $\mathbf{x} \in \mathcal{X}$; $ \mathcal{T} $ is the set of treatments, $T \in \mathcal{T}$; $\mathcal{Y}$ is the set of responses, $Y \in \mathcal{Y}$; $ \mathbbm{t} $ describes the transition probability to 
	$ Y $ (being a terminal state in this bandit setting) from $ \mathbf{x} $ after applying treatment $ T $, thus $ \mathbbm{t}(Y, T, \mathbf{x}) \doteq p(Y | T, \mathbf{x}) $; 
	and $ R $ is the reward function denoted $ R: Y \rightarrow \mathbb{R} $.

	\begin{figure}
		\centering
		\includegraphics[width=.5\columnwidth]{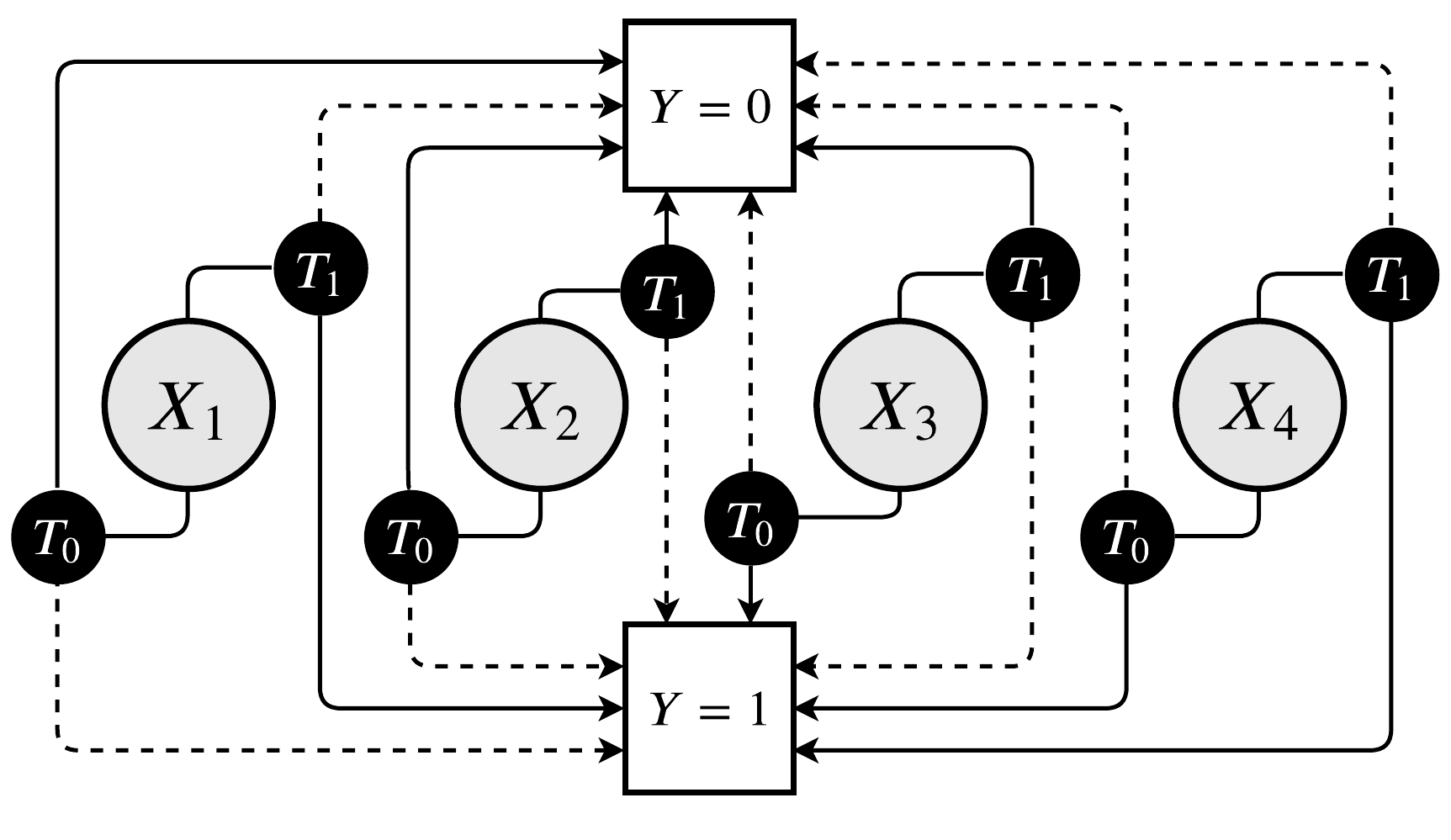}     
		\caption{The MDP for a CMAB in ITE optimisation: grey circles denote individuals of type $ X_j \subset \mathcal{X} $; squares indicate the response $Y=1$ or $Y=0$;
		black circles represent treatments with $T_0$ if $T=0$ and $T_1$ if $T=1$, done so for brevity; dashed arrows are used when $ \mathbbm{t}(Y, T, X_j) = 0 $;
		and full arrows are used when $ \mathbbm{t}(Y, T, X_j) = 1 $.}
		\label{fig:mdpum}
	\end{figure}
	
	As is illustrated in Figure~\ref{fig:mdpum}, we can use this MDP, with $\mathbbm{t} \rightarrow \{0, 1\}$, to subdivide $\mathcal{X}$ into four different kinds of individuals based on their transition properties \cite{devriendt2018}:
	\begin{description}
		\item[$X_1 \subset \mathcal{X}$] Respond ($Y=1$) only when treated ($T=1$)
		\item[$X_2 \subset \mathcal{X}$] Never responds ($Y=0$), regardless of treatment
		\item[$X_3 \subset \mathcal{X}$] Always respond ($Y=1$), regardless of treatment
		\item[$X_4 \subset \mathcal{X}$] Respond ($Y=1$) only when untreated ($T=0$)
	\end{description}

	If $ Y=1 $ is the desired outcome, one can deduct from Figure~\ref{fig:mdpum}, that only individuals from $ X_1 $ 
	yield a positive causal relationship between $ T $ and $ Y $ as applying treatment (i.e., following $ T_1 $) to any 
	other type of individual will either: not result in $ Y=1 $; or will, regardless of $ T $. As an example, take the individuals in $ X_4 $: as both $T=1$ and $T=0$ yield a transition probability of $ \mathbbm{t} = 1 $,
	it does not matter which treatment the agent applies for the individuals to respond ($ Y=1 $). Therefore, a causal agent should only apply treatment ($T=1$) when given an individual from $X_1$.

	Using $\hat{r}$ to differentiate between treatments, an agent would not find an optimum in case of $X_4$. However, adding penalties, $\psi_i$, we can further differentiate between treatments and incorporate $\tau$.

\section{Experiments} \label{sec:results}
	We frame our experiments using the CMAB's objective: 
	(i) ITE prediction (rather than response prediction); and (ii) causal treatment selection.
	As the U-CMAB is a UM method, we compare against the state-of-the art in UM, being an
	uplift random forest (URF) \cite{devriendt2018}.

	\textbf{ITE prediction} is tested using the Hillstrom dataset\footnote{\url{https://blog.minethatdata.com/2008/03/minethatdata-e-mail-analytics-and-data.html}}, a well known resource
	for ITE prediction with two treatments and eighteen variables \cite{devriendt2018,radcliffe2011}. We evaluate performance 
	using a qini-chart (a relative of the gini-chart) \cite{kane2014}: after ranking 
	each individual in a hold-out test-set according to their estimated $\hat{u}$, the 
	cumulative incremental response-rate is calculated using,
	\begin{equation} \label{eq:qini}
		q(b) \doteq \left( \frac{Y_{1, b}}{N_{1, b}} - \frac{Y_{0, b}}{N_{0, b}} \right),
	\end{equation}
	where: $q(b)$ accounts for the first $b \in \mathbb{N}$ bins of size $\frac{N}{B}$; $Y_{i, b}$ is the amount of responders 
	with $T=i$; and $N_{i, b}$ is the amount of individuals treated with $T=i$. As an individual with high $\hat{u}$ is ranked first, (\ref{eq:qini}) should score high for the
	first individuals and gradually decrease when more individuals are included in the evaluation. 

\begin{figure}
		\centering
		\includegraphics[width=.5\columnwidth]{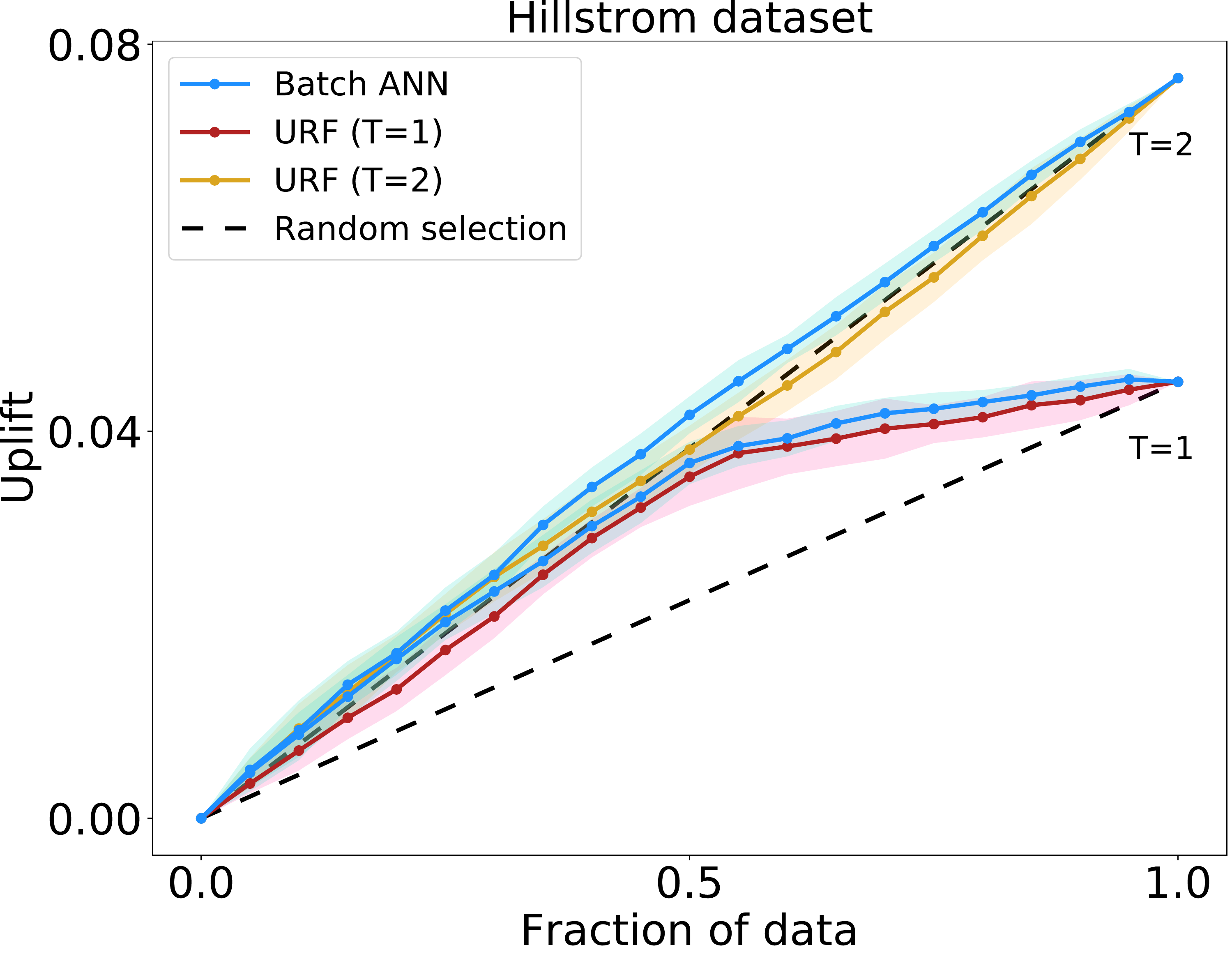}     
		\caption{Compared performance on the Hillstrom dataset of a single batch constrained ANN against two separate URFs, where: URF~(T=1) was trained for treatment $T=1$; and URF~(T=2) was trained for $T=2$. The farther a model is removed from the random selection line, the better.}
		\label{fig:qini}
	\end{figure}

	In our experiment we compared a batch constrained artificial neural network (ANN) \cite{ernst2005,johansson2016}
	to train $\hat{r}_u$, as in (\ref{eq:rew}), against two separate URFs---one for each treatment as current methods can only 
	estimate for one treatment at a time. From Figure~\ref{fig:qini} we recognise that the U-CMAB, using a batch ANN, 
	compares favourably against both URFs, and is thus able to predict the ITE nicely using $\hat{r}_u$.

	\textbf{Causal treatment selection} is tested using a simulated environment \cite{berrevoets2019} allowing us to compare against an all-knowing optimal policy, while controlling how dynamic the environment should be.
	\begin{figure}
		\centering
		\subcaptionbox{No drift\label{fig:exp}}
			{\includegraphics[width=0.3\textwidth]{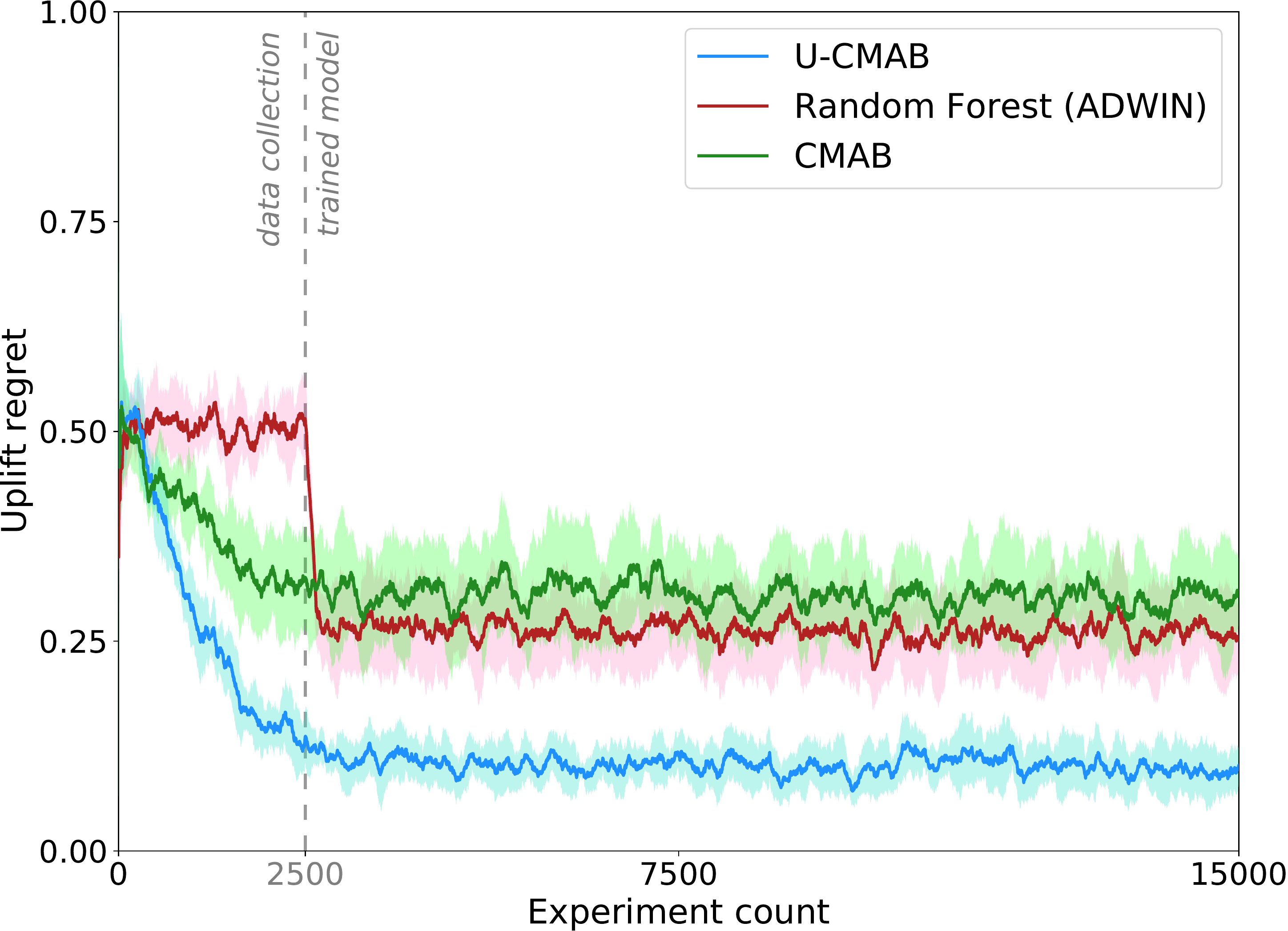}}
		\subcaptionbox{Sudden drift\label{fig:drift1}}
			{\includegraphics[width=0.3\textwidth]{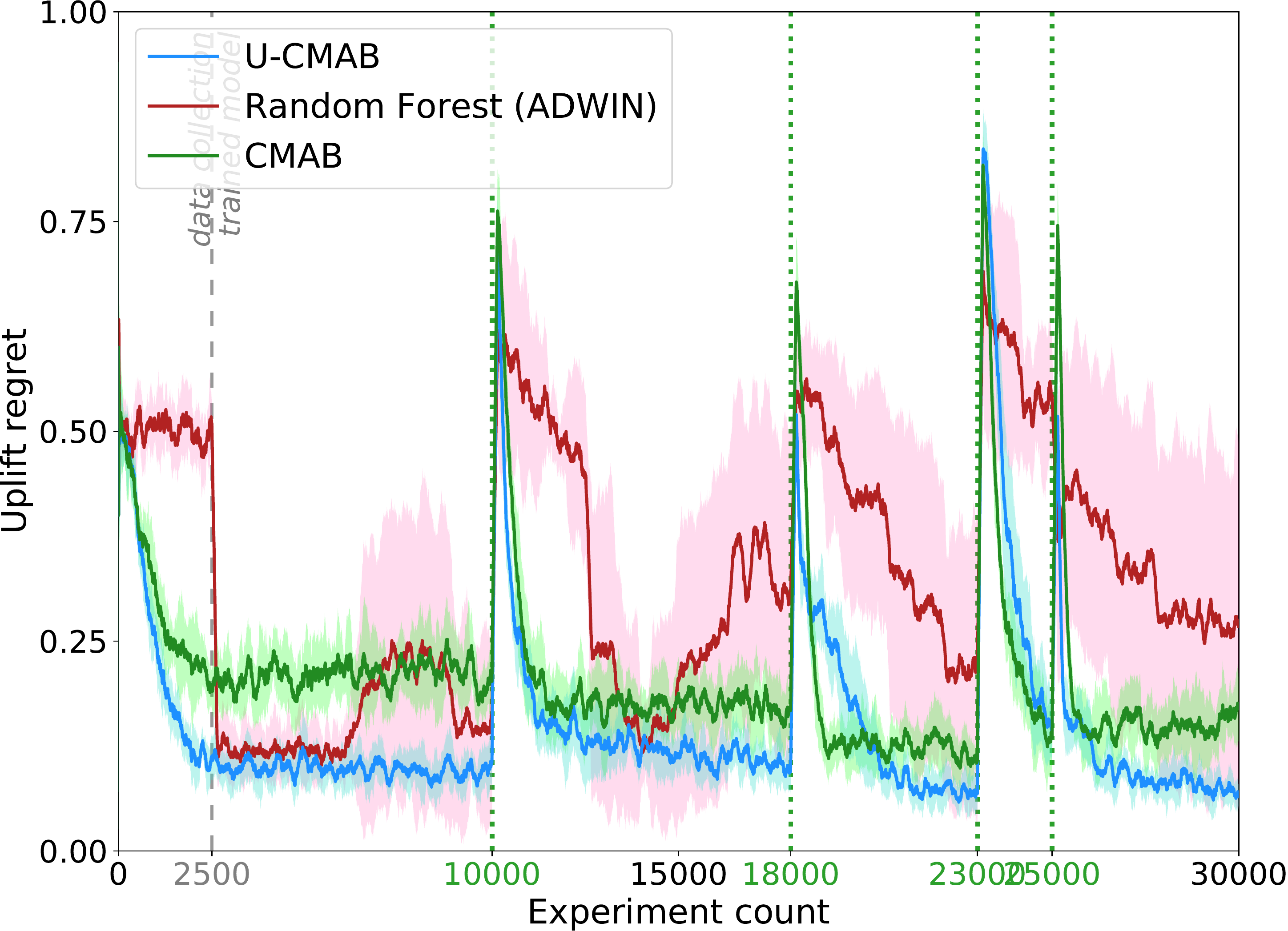}} 
		\subcaptionbox{Gradual drift\label{fig:drift2}}
			{\includegraphics[width=0.3\textwidth]{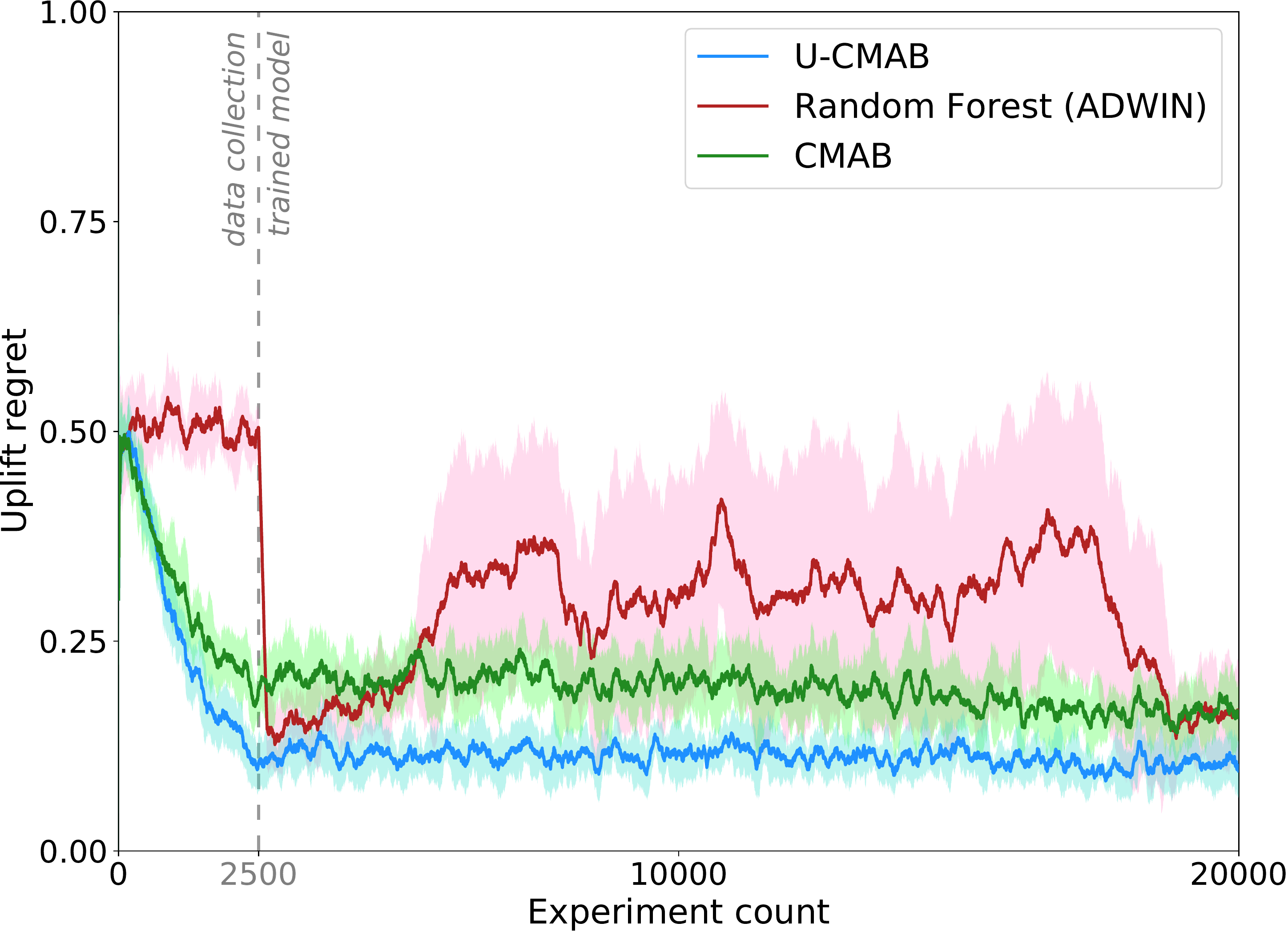}} 
		\caption{Averaged performance over ten runs of the U-CMAB, URF and CMAB in various randomly-generated 
		simulated environments \cite{berrevoets2019}. The grey dashed line indicates 
		the end of the first data gathering period for the URF, yielding a regret of 
		$ 0.5 $ as treatments are applied randomly. Dotted lines in Figure~\ref{fig:drift1} 
		indicate a sudden drift.}\label{fig:results}
	\end{figure} 

	In Figure~\ref{fig:results} we have plotted performance of: (i) a URF \cite{devriendt2018}, which we combined with an adaptive 
	sliding window (ADWIN) change detection algorithm, initiating a data 
	collection and retraining routine when necessary \cite{gama2014}; (ii) a regular 
	CMAB; and (iii) the U-CMAB. We chose an $\epsilon$-greedy training strategy for both bandits for two major reasons: (i) typical implementations use a Robins-Monro estimation 
	of their objective (both $\hat{r}$ and $\hat{r}_u$ are an expectation), which is easily upgraded for dynamic settings using a constant step-size; (ii) $\epsilon$-greedy has been shown to converge in a variety of 
	environments \cite{kuleshov2014} which aids in our setting, as the environment is usually ill-documented \cite{devriendt2018}.
	
	Performance shown is measured in a regret metric, taking into 
	account the causal nature of each treatment decision \citep{berrevoets2019}. Our results clearly indicate a performance increase in both dynamic and static environments, while confirming immense instability of the URF in dynamic environments, even when ameliorated with an ADWIN change detection strategy. As expected, the CMAB performs worst in a static environment (Figure~\ref{fig:exp}) since it is the only method not optimising an ITE, however, it outperforms the URF in dynamic environments (Figures~\ref{fig:drift1}~and~\ref{fig:drift2}) further confirming the importance of dynamic methods.

\section{Conclusion}
	Through the results shown in Section~\ref{sec:results}, we provide 
	evidence that (\ref{eq:tau}) and (\ref{eq:ucmab}) allow bandit algorithms to make treatment
	decisions based on a prediction for the individual-treatment-effect. The use of bandits 
	minimises the amount of random experiments through efficient exploration
	and offers resilience against a dynamic environment.

	In light of further work, we are interested in the U-CMAB's extension to full
	reinforcement learning \cite{sutton2018} using an estimated $\tau$ through time, potentially allowing an agent to make causal decisions 
	leading to more efficient use of resources. Efficiently managing resources 
	required to obtain a certain reward could greatly affect the application in 
	practical settings.

  \def\bibfont{\footnotesize}
  \setlength{\bibsep}{2pt}

\begin{thebibliography}{18}
\makeatletter
\newcommand{\dinatlabel}[1]%
{\ifNAT@numbers\else\NAT@biblabelnum{#1}\hspace{2\labelsep}\fi}
\makeatother
\expandafter\ifx\csname natexlab\endcsname\relax\def\natexlab#1{#1}\fi
\expandafter\ifx\csname url\endcsname\relax\def\url#1{\texttt{#1}}\fi

\bibitem[Berrevoets und Verbeke(2019)]{berrevoets2019}
\dinatlabel{Berrevoets und Verbeke 2019} \textsc{Berrevoets}, Jeroen~;
  \textsc{Verbeke}, Wouter:
\newblock Causal Simulations for Uplift Modeling.
\newblock In: \emph{arXiv preprint arXiv:1902.00287}
\newblock (2019)

\bibitem[Devriendt u.\,a.(2018)Devriendt, Moldovan und Verbeke]{devriendt2018}
\dinatlabel{Devriendt u.\,a. 2018} \textsc{Devriendt}, Floris~;
  \textsc{Moldovan}, Darie~; \textsc{Verbeke}, Wouter:
\newblock A Literature Survey and Experimental Evaluation of the
  State-of-the-Art in Uplift Modeling: A Stepping Stone Toward the Development
  of Prescriptive Analytics.
\newblock In: \emph{Big Data}
\newblock 6 (2018), Nr.~1, S.~13--41. --
\newblock URL \url{https://doi.org/10.1089/big.2017.0104}. --
\newblock PMID: 29570415

\bibitem[Ernst u.\,a.(2005)Ernst, Geurts und Wehenkel]{ernst2005}
\dinatlabel{Ernst u.\,a. 2005} \textsc{Ernst}, Damien~; \textsc{Geurts},
  Pierre~; \textsc{Wehenkel}, Louis:
\newblock Tree-based batch mode reinforcement learning.
\newblock In: \emph{Journal of Machine Learning Research}
\newblock 6 (2005), Nr.~Apr, S.~503--556

\bibitem[Fang(2018)]{fang2018}
\dinatlabel{Fang 2018} \textsc{Fang}, Xiao:
\newblock \emph{Uplift Modeling for Randomized Experiments and Observational
  Studies}, Massachusetts Institute of Technology, Dissertation, 2018

\bibitem[Gama u.\,a.(2014)Gama, {\v{Z}}liobait{\.e}, Bifet, Pechenizkiy und
  Bouchachia]{gama2014}
\dinatlabel{Gama u.\,a. 2014} \textsc{Gama}, Jo{\~a}o~;
  \textsc{{\v{Z}}liobait{\.e}}, Indr{\.e}~; \textsc{Bifet}, Albert~;
  \textsc{Pechenizkiy}, Mykola~; \textsc{Bouchachia}, Abdelhamid:
\newblock A survey on concept drift adaptation.
\newblock In: \emph{ACM computing surveys (CSUR)}
\newblock 46 (2014), Nr.~4, S.~44

\bibitem[Gutierrez und G{\'e}rardy(2017)]{gutierrez2017}
\dinatlabel{Gutierrez und G{\'e}rardy 2017} \textsc{Gutierrez}, Pierre~;
  \textsc{G{\'e}rardy}, Jean-Yves:
\newblock Causal Inference and Uplift Modelling: A Review of the Literature.
\newblock In: \emph{International Conference on Predictive Applications and
  APIs}, 2017, S.~1--13

\bibitem[Johansson u.\,a.(2016)Johansson, Shalit und Sontag]{johansson2016}
\dinatlabel{Johansson u.\,a. 2016} \textsc{Johansson}, Fredrik~;
  \textsc{Shalit}, Uri~; \textsc{Sontag}, David:
\newblock Learning representations for counterfactual inference.
\newblock In: \emph{International conference on machine learning}, 2016,
  S.~3020--3029

\bibitem[Kane u.\,a.(2014)Kane, Lo und Zheng]{kane2014}
\dinatlabel{Kane u.\,a. 2014} \textsc{Kane}, Kathleen~; \textsc{Lo},
  Victor~S.~; \textsc{Zheng}, Jane:
\newblock Mining for the truly responsive customers and prospects using
  true-lift modeling: Comparison of new and existing methods.
\newblock In: \emph{Journal of Marketing Analytics}
\newblock 2 (2014), Nr.~4, S.~218--238

\bibitem[Kuleshov und Precup(2014)]{kuleshov2014}
\dinatlabel{Kuleshov und Precup 2014} \textsc{Kuleshov}, Volodymyr~;
  \textsc{Precup}, Doina:
\newblock Algorithms for multi-armed bandit problems.
\newblock In: \emph{arXiv preprint arXiv:1402.6028}
\newblock (2014)

\bibitem[Pearl(2009)]{pearl2009}
\dinatlabel{Pearl 2009} \textsc{Pearl}, Judea:
\newblock \emph{Causality}.
\newblock Cambridge, UK~: Cambridge university press, 2009

\bibitem[Radcliffe und Surry(2011)]{radcliffe2011}
\dinatlabel{Radcliffe und Surry 2011} \textsc{Radcliffe}, Nicholas~J.~;
  \textsc{Surry}, Patrick~D.:
\newblock Real-world uplift modelling with significance-based uplift trees.
\newblock In: \emph{White Paper TR-2011-1, Stochastic Solutions}
\newblock (2011)

\bibitem[Robbins(1952)]{robbins1952}
\dinatlabel{Robbins 1952} \textsc{Robbins}, Herbert:
\newblock Some aspects of the sequential design of experiments.
\newblock In: \emph{Bulletin of the American Mathematical Society}
\newblock 55 (1952), S.~527--535

\bibitem[Rubin(2005)]{rubin2005}
\dinatlabel{Rubin 2005} \textsc{Rubin}, Donald~B.:
\newblock Causal Inference Using Potential Outcomes.
\newblock In: \emph{Journal of the American Statistical A}
\newblock 100 (2005), Nr.~469, S.~322--331. --
\newblock URL \url{https://doi.org/10.1198/016214504000001880}

\bibitem[Saffari u.\,a.(2009)Saffari, Leistner, Santner, Godec und
  Bischof]{saffari2009}
\dinatlabel{Saffari u.\,a. 2009} \textsc{Saffari}, Amir~; \textsc{Leistner},
  Christian~; \textsc{Santner}, Jakob~; \textsc{Godec}, Martin~;
  \textsc{Bischof}, Horst:
\newblock On-line random forests.
\newblock In: \emph{2009 ieee 12th international conference on computer vision
  workshops, iccv workshops}
\newblock IEEE (Veranst.), 2009, S.~1393--1400

\bibitem[Shalit u.\,a.(2017)Shalit, Johansson und Sontag]{shalit2017}
\dinatlabel{Shalit u.\,a. 2017} \textsc{Shalit}, Uri~; \textsc{Johansson},
  Fredrik~D.~; \textsc{Sontag}, David:
\newblock Estimating individual treatment effect: generalization bounds and
  algorithms.
\newblock In: \emph{Proceedings of the 34th International Conference on Machine
  Learning-Volume 70}
\newblock JMLR. org (Veranst.), 2017, S.~3076--3085

\bibitem[Sutton und Barto(2018)]{sutton2018}
\dinatlabel{Sutton und Barto 2018} \textsc{Sutton}, Richard~S.~;
  \textsc{Barto}, Andrew~G.:
\newblock \emph{Reinforcement learning: An introduction}.
\newblock 2nd.
\newblock Cambridge, MA, USA~: MIT press, 2018

\bibitem[Tsymbal(2004)]{tsymbal2004}
\dinatlabel{Tsymbal 2004} \textsc{Tsymbal}, Alexey:
\newblock The problem of concept drift: definitions and related work
\newblock / Computer Science Department, Trinity College Dublin.
\newblock Citeseer, 2004.
\newblock  -- Forschungsbericht

\bibitem[Zhou(2015)]{zhou2015}
\dinatlabel{Zhou 2015} \textsc{Zhou}, Li:
\newblock A survey on contextual multi-armed bandits.
\newblock In: \emph{arXiv preprint arXiv:1508.03326}
\newblock (2015)

\end{thebibliography}

  \newpage
  \section{Appendix}
	\subsection{Reproducibility}
	Python code used to test the U-CMAB as in Section~\ref{sec:results} is provided online
	\url{https://github.com/vub-dl/u-cmab}. In this code you will find hyperparameters,
	notebooks documenting plot methods and extra visualisations and experiments further 
	confirming current instability. 
	
	\subsection{Proof of (\ref{eq:ucmab})} \label{ap:proof}
	\begin{proof}
		We prove that the equality,
		\begin{equation*}
			\tau = \frac{\psi_1 - \psi_0}{R(Y=1)},
		\end{equation*}
		allows a bandit to make decisions based on some $\tau$ as in (\ref{eq:tau}). 
		By introducing a penalty $ \psi_i $ of a 
		treatment $ T=i $ in the treatment selection procedure
		as in (\ref{eq:bellman}) and (\ref{eq:rew}), 
		\begin{equation} \label{eq:dec}
			T^* = \argmax_i \left\{\mathbb{E}[R(Y) - \psi_i | T=i, \mathbf{x}]\right\},
		\end{equation}
		reflecting the definition of $ \hat{r}_u $.
		In case of a single treatment ($ T=1 $) and control ($ T=0 $), the 
		$ \argmax_i\{\cdot\} $ in (\ref{eq:dec}) can be simplified in,
		\begin{equation} \label{eq:ineq}
			T^* = \mathbb{I}[\hat{r}(T=1, \mathbf{x}) - \psi_1 > \hat{r}(T=0, \mathbf{x}) - \psi_0],
		\end{equation}
		as $\psi_i$ is a constant and $\mathbb{E}[\cdot]$ a linear operator,
		with $ \hat{r} $ as an expected value based on the transition function
		\cite{sutton2018},
		\begin{equation} \label{eq:rew_exp}
			\hat{r}(T, \mathbf{x}) \doteq R(Y=1)\hat{p}(Y=1 | T, \mathbf{x}),
		\end{equation}
		with $ R(Y=1) $ as the reward received after responding to $ T $.
	
		After rearranging (\ref{eq:rew_exp}) into (\ref{eq:ineq}) we get,
		\begin{equation} \label{eq:rear}
			T^* = \mathbb{I}[R(Y=1)\hat{p}(Y=1 | T=1, \mathbf{x}) - \psi_1 > R(Y=1)\hat{p}(Y=1 | T=0, \mathbf{x}) - \psi_0].
		\end{equation}
		
		Rearranging (\ref{eq:rear}) yields,
		\begin{equation}
			T^* = \mathbb{I}\left[\hat{u}(Y, T, \mathbf{x}) > \frac{\psi_1 - \psi_0}{R(Y=1)}\right],
		\end{equation}
		which through (\ref{eq:tau}) implies,
		\begin{equation*}
			\tau = \frac{\psi_1 - \psi_0}{R(Y=1)}
		\end{equation*}
	\end{proof}

  \end{document}